\documentclass{article}

\usepackage{arxiv}
\usepackage{float}
\usepackage[utf8]{inputenc} % allow utf-8 input
\usepackage[T1]{fontenc}    % use 8-bit T1 fonts
\usepackage{hyperref}       % hyperlinks
\usepackage{url}            % simple URL typesetting
\usepackage{booktabs}       % professional-quality tables
\usepackage{amsfonts}       % blackboard math symbols
\usepackage{nicefrac}       % compact symbols for 1/2, etc.
\usepackage{microtype}      % microtypography
\usepackage{cleveref}       % smart cross-referencing
\usepackage{lipsum}         % Can be removed after putting your text content
\usepackage{graphicx}
\usepackage{natbib}
\usepackage{doi}
\usepackage{subfigure}
\usepackage{lineno}
\usepackage{color}
\usepackage{multirow}
\usepackage{graphicx}
\usepackage{amssymb}
\usepackage{fontawesome}

\title{RCDT:Relational Remote Sensing Change Detection with Transformer}

\date{September 9, 2022}
\author{ {\hspace{1mm}Kaixuan Lu} \\
    Aerospace Information Research Institute\\
	Chinese Academy of Sciences\\
	Beijing 100094, China \\
	\And
	{\hspace{1mm}Xiao Huang} \\
	Department of Geosciences\\
	University of Arkansas\\
	Fayetteville, Arkansas 72701, USA\\
}
\begin{document}
\maketitle

\begin{abstract}
Deep learning based change detection methods have received wide attentoion, thanks to their strong capability in obtaining rich features from images. However, existing AI-based CD methods largely rely on three functionality-enhancing modules, i.e., semantic enhancement, attention mechanisms, and correspondence enhancement. The stacking of these modules leads to great model complexity. To unify these three modules into a simple pipeline, we introduce Relational Change Detection Transformer (RCDT), a novel and simple framework for remote sensing change detection tasks. The proposed RCDT consists of three major components, a weight-sharing Siamese Backbone to obtain bi-temporal features, a Relational Cross Attention Module (RCAM) that implements offset cross attention to obtain bi-temporal relation-aware features, and a Features Constrain Module (FCM) to achieve the final refined predictions with high-resolution constraints. Extensive experiments on four different publically available datasets suggest that our proposed RCDT exhibits superior change detection performance compared with other competing methods. The therotical, methodogical, and experimental knowledge of this study is expected to benefit future change detection efforts that involve the cross attention mechanism.
\end{abstract}

% keywords can be removed

\section{Introduction}
\label{S:1}
With the rapid development of remote sensing observational techniques, change detection (CD) \citep{singh1989review} that relies on multi-temporal very high resolution (VHR) images has received wide attention, given its large potential in a variety of domains that include urban management, agricultural monitoring, disaster relief, and environmental monitoring. Imagery-based CD aims to detect the locations and categories of changed pixels in multi-temporal images. As one of the most important tasks in the remote sensing interpretation domain, many challenges in CD still remain, such as the limitation of short-range semantic information and the loss of bi-temporal feature correspondence.

Existing CD methods can be grouped into two major categories, i.e., traditional CD approaches and AI-based CD approaches. Traditional CD approaches include pixel-based CD methods that focus on spectral information and object-based CD methods that focus on spatial patterns \citep{hussain2013change}. Pixel-based CD methods tend to establish decision functions, aiming to derive optimal thresholds for determining whether the corresponding pixels of bi-temporal images change. Notable pixel-based CD methods include Image differencing \citep{coops2007identifying}, Image ratioing \citep{howarth1981procedures}, Regression analysis \citep{ludeke1990analysis}, Vegetation index differencing \citep{nordberg2005vegetation}, Change vector analysis (CVA) \citep{bayarjargal2006comparative}, Principal component analysis (PCA) \citep{deng2008pca}, and Tasselled cap transformation (KT) \citep{jin2005comparison}. Despite their wide applications, pixel-based CD methods are only preferred when dealing with multispectral or hyperspectral images. In addition, their performances largely depend on the decision functions and threshold settings. In comparison, object-based CD methods \citep{addink2012introduction,tomowski2011colour,cai2013comparison} rely on image segmentation or stratification techniques to extract meaningful image objects, which serve as analytical units for subsequent change analysis. Different from pixel-based CD methods, object-based CD methods are able to take advantage of various properties of an image, such as the spectral, textual, spatial, and topological information, as well as the hierarchical object characteristics. However, the accuracy and consistency of segmentation of object-based CD methods have been criticized by many as under- and/or over-segmentation errors tend to occur. Traditional CD approaches, largely relying on hand-craft features, have limited capability in describing and capturing complex changes, leading to their tendency to overfitting and unsatisfactory generalizability. AI-based CD approaches, however, rely on multilayer deep learning architectures \citep{lecun2015deep}. Such a design facilitates the learning of complex mapping mechanisms, which ensures improved (both in accuracy and efficiency) change detection results. In this study, we aim to advance AI-based CD approaches for VHR images by establishing a novel paradigm featured by its high performance, efficiency and simplicity.

AI-based CD approaches started to emerge and gradually became a research hotspot in recent years, thanks to the advances in computation and the availability of a massive number of high-resolution image datasets. The pioneering effort, i.e., FCS \citep{alcantarilla2018street,daudt2018fully}, implemented end-to-end fully convolutional Siamese architecture to extract bi-temporal features, followed by concatenation or differencing fusion operations before the decoder. However, the fusion of bi-temporal features in the encoder fails to fully explore the correspondence of bi-temporal features, leading to insufficient multi-scale semantic information. To enrich multi-scale, long-range semantic features, a variety of methods \citep{lei2019landslide,zheng2021clnet,fang2021snunet} adopted structures, such as SPP \citep{he2015spatial} and Unet \citep{ronneberger2015u}, for extracting bi-temporal features. To obtain richer bi-temporal feature correspondence, scholars \citep{zhang2020deeply,ding2021dsa,liu2022end} started to design bi-temporal feature fusion modules or domain-adaptative modules to enhance correspondence by filtering out background noises. To enhance models’ sensitivity to changes, scholars implemented spatial and channel attention to direct models’ focus to potential changed areas \citep{song2021suacdnet,chen2020spatial,shi2021deeply}. These attention mechanisms greatly benefit the distinguishment of foreground and background. Despite the great success of AI-based CD approaches in VHR images, most of them tend to rely on convolutional operations, failing to take full advantage of the long-range and global semantic information. In addition, most AI-based CD approaches essentially build bi-temporal feature correspondence via linear processes between bi-temporal features. The trained models tend to present satisfactory performance on specific training datasets where the models can well converge but fail to adapt to random, complicated, or noise-prone scenarios. The overlay of modules for semantic enhancement, correspondence enhancement, and attention mechanism makes these models complicated, computationally demanding, and difficult to converge. Thus, it is necessary to simplify existing CD pipelines without sacrificing CD accuracy.

In this work, we propose Relational Change Detection Transformer (RCDT), a novel CD paradigm that achieves efficient and highly accurate CD for VHR images. The proposed RCDT obtains bi-temporal features via a weight-sharing Siamese Backbone. To collect contextual information at varying positions, we design a light FPN \citep{lin2017feature} to obtain multi-scale semantic features. The proposed RCDT is featured by a designed Relational Cross Attention Module (RCAM) that obtains relational features among bi-temporal images. The proposed RCAM adopts an offset-cross attention operation to derive relational features between bi-temporal features and implements standard cross attention and Feed-Forward Network (FFN) to derive pixel embeddings with rich global information. To improve the detection of small areas in VHR images, we propose a Features Constrain Module (FCM), an efficient strategy that enhances pixel embedding via high-resolution feature constraints. We evaluate the proposed RCDT on four popular CD datasets that include LEVIR-CD \citep{chen2020spatial}, DSIFN \citep{zhang2020deeply}, CDD \citep{lebedev2018change} and SYSU-CD \citep{shi2021deeply}. The results prove the superior performance of the proposed RCDT over other competing AI-based CD methods.

\section{Related work}
\label{S:2}
\subsection{AI-Based Change Detection}
\label{S:2.1}
Existing AI-based approaches tend to follow the Siamese network paradigm, which usually consists of two weight-sharing CNN branches that benefit the projection of bi-temporal inputs into the same latent space. Contextual feature fusion and spatial/channel attention are two widely used techniques in these approaches. \citep{daudt2018fully} proposed a fully convolutional Siamese architecture to collect bi-temporal features, where bi-temporal features were fused via concatenation operation or difference operation, serving as the inputs to the decoder. \citep{fang2021snunet} designed a densely connected Siamese network that maintains high-resolution and fine-grained bi-temporal features for change detection. \citep{chen2022fccdn} took advantage of two densely connected branches to fuse bi-temporal features by ensembling multistage features. \citep{liu2022end} proposed a domain adaptation framework, aiming to reduce the background disparities in bi-temporal features. \citep{song2021suacdnet,chen2020spatial,shi2021deeply,shen2022semantic} developed spatial, channel, or hybrid attention mechanisms to enhance significant features by suppressing unimportant features. Most exiting AI-based CD approaches seek solutions from three perspectives, i.e., harnessing the rich semantic information, exploring the correspondence of bi-temporal features, and relying on attention mechanisms. Despite the successes of the aforementioned methods in CD takes, challenges still remain. These methods tend to build bi-temporal feature correspondence via linear processes. Such a strategy often leads to satisfactory performance on specific training sets while failing to adapt to complex, noise-prone scenes or scenes that are considerably different from the training set. In addition, the bulkiness of models with overlapping functionality-enhancing modules (e.g., semantic enhancement, attention mechanisms, and correspondence enhancement) unavoidably results in additional difficulty in model convergence and increased computational demand. In this study, we seek to resolve these issues by proposing a simplified pipeline.

\subsection{Vision Transformer}
\label{S:2.2}
Transformer was first proposed by \citep{vaswani2017attention} as a new building block based solely on attention mechanisms for machine translation. Recent years have seen its rapid development, as many efforts have been made to apply Transformers in computer vision tasks \citep{dosovitskiy2020image,carion2020end,liu2021swin,zhu2020deformable,strudel2021segmenter,zheng2021rethinking} and received great success due to its strong capability in long-range feature expression. These methods first decomposed an image into multiple patches, forming a sequence fed to Transformer architectures, where each element of a sequence is able to aggregate global information. In the CD domain, \citep{chen2021remote} adopted a Siamese Transformer encoder-decoder structure to enhance bi-temporal feature expression and used a difference operation to fuse bi-temporal features. \citep{bandara2022transformer} designed a Siamese Hierarchical Transformer as encoder and an FFN as decoder, where bi-temporal features from each phase of the encoder were fused via the difference operation. The hierarchical fused features were further concatenated and fed to the decoder. \citep{liu2022cnn} developed multi-scale Transformer encoders and decoders to aggregate multi-scale contextual information and further concatenated them to multi-branch prediction heads. \citep{shi2022divided} proposed two cascaded self-attention modules to capture feature correspondence in an early-fusion multi-temporal image. Although these methods adopted Transformer structures to enrich contextual information of bi-temporal features, they fail to harness the inner relations of bi-temporal features, which is crucial in CD tasks. We argue that the essence of CD tasks is not about the separate characteristics in bi-temporal features but the relation between bi-temporal features. The cross attention mechanism in Transformers offers correlation modeling with a global receptive field, making it an ideal strategy to solve remote sensing CD tasks. In this study, we aim to simplify existing AI-based CD pipelines by proposing a novel and simple framework that relies on the cross attention mechanism in Transformers. The proposed architecture is able to excavate the long-range contextual information and correspondence between bi-temporal inputs.

\section{Methodology}
\label{S:3}
In this section, we introduce the architecture of RCDT (Section \ref{S:3.1}), the Siamese backbone for feature extraction (Section \ref{S:3.2}), the transformer-based Relational Cross Attention Module (RACM) (Section \ref{S:3.3}), the Features Constrain Module (FCM) (Section \ref{S:3.4}), and the loss function (Section \ref{S:3.5}).
\subsection{Overview}
\label{S:3.1}
The primary goal of CD is to identify changed areas by interpreting bi-temporal images captured in the same geographic area. Given $I^{T_{1}} \in \mathbb{R} ^{H\times W\times 3}$ and $I^{T_{2}} \in \mathbb{R} ^{H\times W\times 3}$ as two bi-temporal images with three channels, a deep learning model aims to predict the probability distributed over all possible  categories for every corresponding pixel in a detected image-pair: $C \in \mathbb{R} ^{K\times H\times W}$, where $H$, $W$ and $K$ denote the image height, image width, and the dual-pixel category, respectively. For Binary Change Detection (BCD), $K\in \left \{ 0,1 \right \}$, with 1 and 0 denoting changed and unchanged.

\begin{figure}[]
	\centering
	\vspace{-0.8cm}
	\includegraphics[width=1\linewidth]{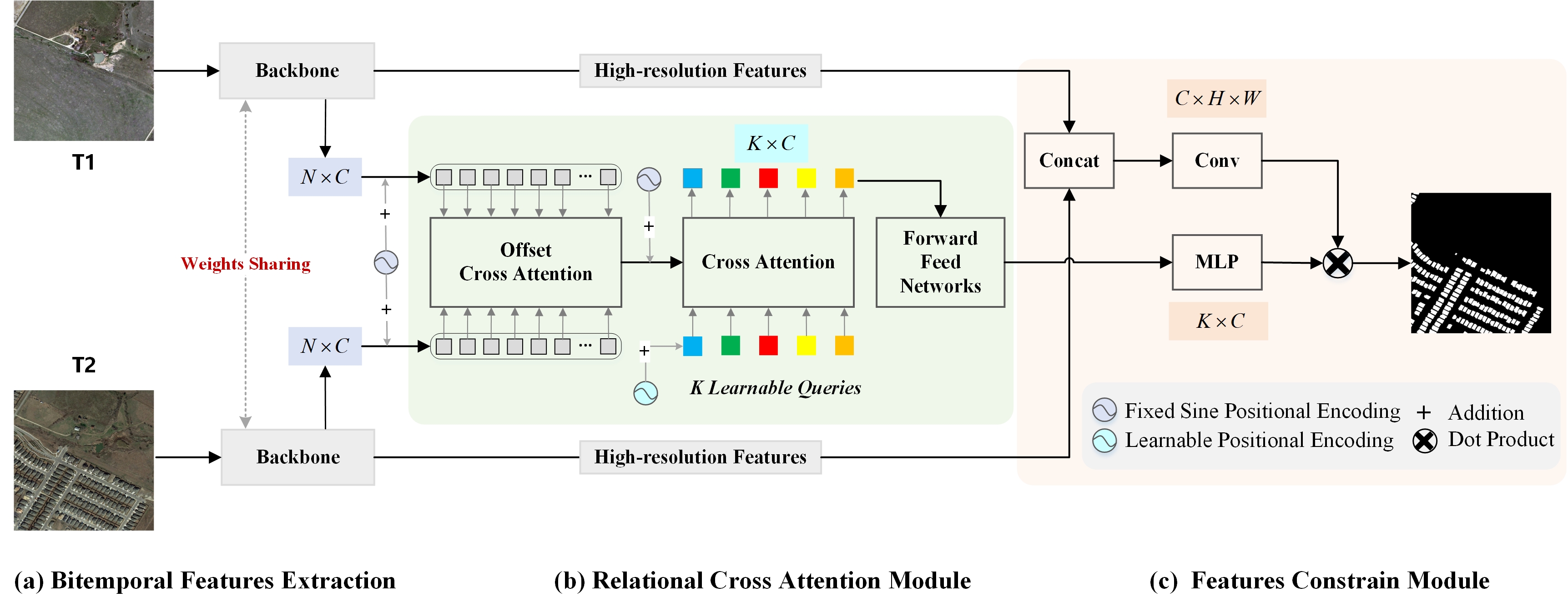}
	\caption{The overall structure of the proposed RCDT.}
	\setlength{\belowcaptionskip}{-2cm}
	\label{fig:1}
\end{figure}

The RCDT we proposed is a novel change detection framework featured by its simplicity, as shown in Figure \ref{fig:1}, which contains three major components, i.e., the Siamese Backbone, the Relational Cross Attention Module (RCAM), and the Feature Constrain Module (FCM):

\textbf{Siamese Backbone.} Following previous methods \citep{daudt2018fully,fang2021snunet}, we adopt the Siamese structure as the backbone to obtain multi-scale features from bi-temporal images. The adopted Siamese Backbone consists of two weight-sharing encoder-decoders whose primary goal is to project and align bi-temporal image features in the same latent space, which benefits subsequent processes that leverage relation-aware feature difference.

\textbf{Relational Cross Attention Module (RCAM).} Inspired by \citep{carion2020end,zhu2020deformable,zhou2022pttr,cheng2022masked}, we propose RCAM, a novel attention mechanism that is different from existing approaches relying on heavy convolution operations to aggregate bi-temporal features. RCAM follows the decoder design in the Transformer, consisting of two cross attention submodules and one FFN submodule (all submodules work in parallel). The first cross attention submodule obtains important bi-temporal relation-aware features while the latter cross attention and FNN submodules capture context-rich pixel embeddings with global information.

\textbf{Features Constrain Module (FCM).} The feature maps from RCAM can be entered into a prediction head that consists of $1\times 1$ convolution to obtain the prediction with a shape of $K\times H\times W$. However, such a procedure inevitably leads to positional information loss, which demands constraint enhancement via high-resolution features. Relying on the output from RCAM, The FCM we proposed obtains the refined predictions ($K\times H\times W$) with high-resolution constraint information by first acquiring pixel embeddings ($K\times C$) via a Multi-Layer Perceptron (MLP), concatenating bi-temporal high-resolution features, and further feeding them to a $1\times 1$ convolutional block. After the dot product operation, the output shape of FCM is $K\times H\times W$. 
\subsection{Siamese Backbone}
\label{S:3.2}
\begin{figure}[H]
	\centering
	\vspace{-0.8cm}
	\includegraphics[width=0.5\linewidth]{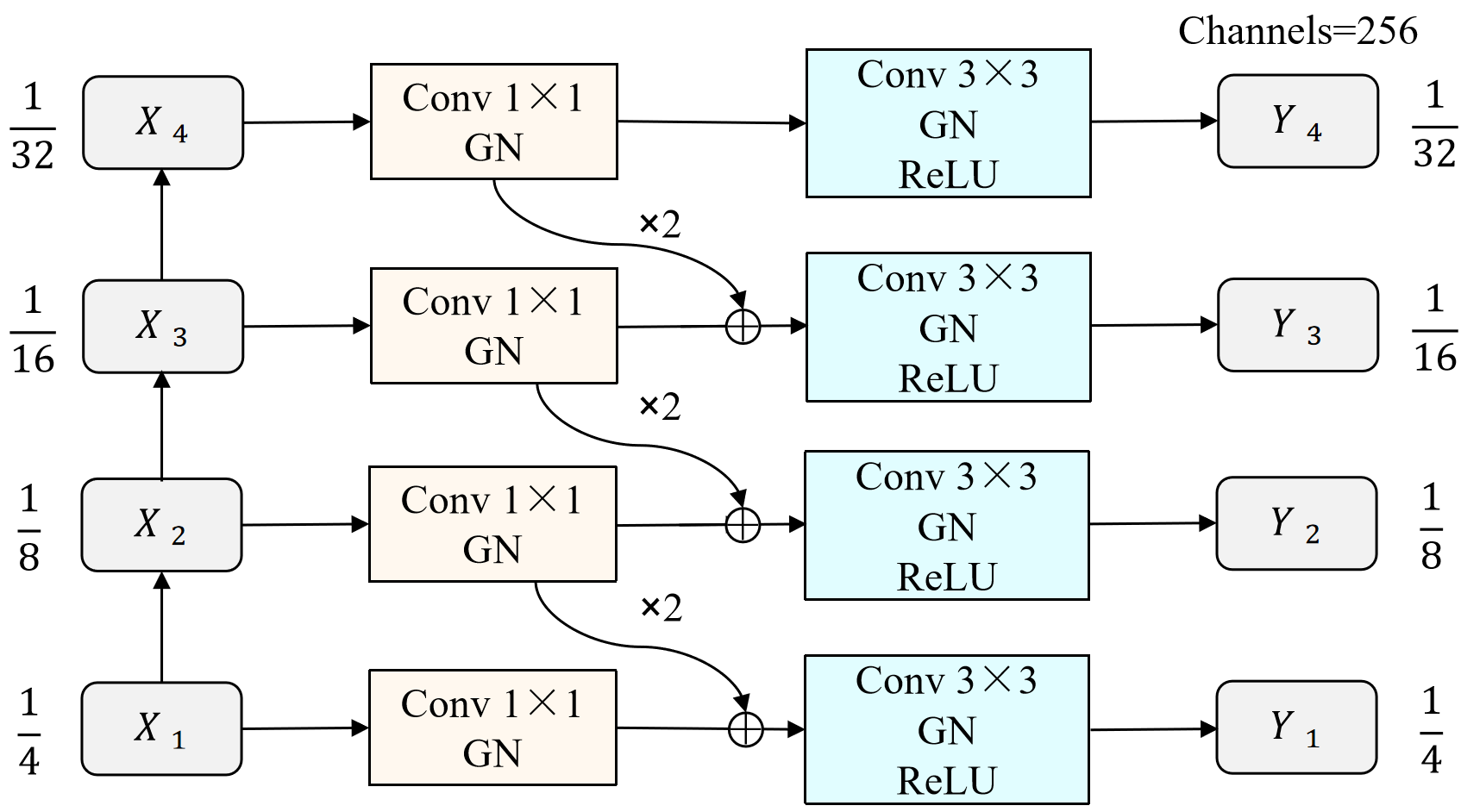}
	\caption{A light FPN structure.}
	\setlength{\belowcaptionskip}{-2cm}
	\label{fig:3}
	\end{figure}

The Siamese Backbone follows an encoder-decoder architecture, with two encoder and decoder sharing weights to ensure that the bi-temporal features are aligned in the latent space. The encoder extracts features from RGB input images ($H\times W\times 3$) and generates low-resolution feature maps $\mathcal{F}  \in \mathbb{R} ^{C_{\mathcal{F}} \times \frac{H}{32} \times \frac{W}{32}} $ in a multi-stage process with varying step sizes ($S=[4,8,16,32]$), where $C_{\mathcal{F}}$ is the number of channels designed by the network. Further, the decoder gradually upsamples low-resolution feature maps to the corresponding high-resolution ones. Note that the adopted encoder-decoder architecture can be replaced by other popular semantic segmentation architectures, such as Unet \citep{ronneberger2015u}, DeepLab \citep{chen2017deeplab}, and PSP \citep{zhao2017pyramid}, to list a few. In light of the strong capability in obtaining the long-range semantic context of the Transformer structure and the necessity to reduce network complexity, we adopt FPN \citep{lin2017feature}, a relatively light structure as our decoder, as shown in Figure \ref{fig:3}. We first use a $1\times 1$ convolution to project the multi-scale feature maps, i.e., $X_{1}$,$X_{2}$,$X_{3}$,$X_{4}$, to a unified channel number (256). Further, we gradually upsample multi-scale feature maps via a bilinear function from the lowest resolution level and add the upsampled feature maps with neighboring feature maps. We then implement convolutional operations that include a $3\times 3$ kernel size, group norm \citep{wu2018group}

\subsection{Relational Cross Attention Module (RCAM)}
\label{S:3.3}
Context has been proved to be essential in extracting image features. Existing change detection models tend to fuse contextual information to enhance the correspondence of features in bi-temporal images. Inspired by the success of Transformer, especially the global attention mechanism that captures long-range semantics information, we aim to explore the potential of attention mechanisms in capturing bi-temporal feature correspondence. Although BIT \citep{chen2021remote}, ChangeFormer \citep{bandara2022transformer} and other Transformer-based CD methods \citep{liu2022cnn,shi2022divided} adopted a Transformer architecture for CD, the correlations of bi-temporal features in these methods were still not fully harnessed, while the major goal of the Transformer was to enhance bi-temporal feature expression. A standard transformer decoder \citep{vaswani2017attention} includes three components, i.e., a self-attention module, a cross attention module, and a feed-forward network(FFN). In DETR \citep{carion2020end}, sine spatial positional encoding is added to queries and keys at every multi-head self-attention layer. 

To reduce the computational demand and facilitate fast convergence, we directly adopt the cross attention module while discarding the self-attention module, as the nature of change detection tasks is to compare bi-temporal information; therefore, it is cross attention driven. First, we project the output bi-temporal features from the Siamese Backbone to a two-dimensional space, as Transformers demand the input of sequential sets. That is to say, the proposed RCAM flattens and permutes the features $\mathcal{F}  \in \mathbb{R} ^{C \times \frac{H}{S} \times \frac{W}{S}}$ to features $\mathcal{F}  \in \mathbb{R} ^{\frac{H\cdot W}{S^{2} } \times C}$, where $C=256$ and $S$ denotes the stride.

To gather cross-contextual information between bi-temporal features, features from the “before” image, i.e., $X_{1}$, serve as input vectors “Query”, while features from the “after” image, i.e., $X_{2}$, serve as “Key” and “Value” with fixed sine spatial positional encoding, as shown in Figure \ref{fig:4}. Different from traditional cross attention that calculates the dot product of “Query” and “Key”, we first derive normalized query and key features via L2 norm and calculate their cosine similarity, aiming to enhance bi-temporal relation attention information. The relation attention map is later normalized with a Softmax operation, and further joins the dot product with value features. To mitigate the influence of noises, instead of deriving the final attention map by adding the previously normalized attention map with query features like ResNet \citep{he2016deep}, we adopt the subtraction operation. The standard cross attention follows:

\begin{figure}[]
	\centering
	\vspace{-0.8cm}
	\subfigure[A demonstration of a standard cross attention]{
		\includegraphics[width=0.5\linewidth]{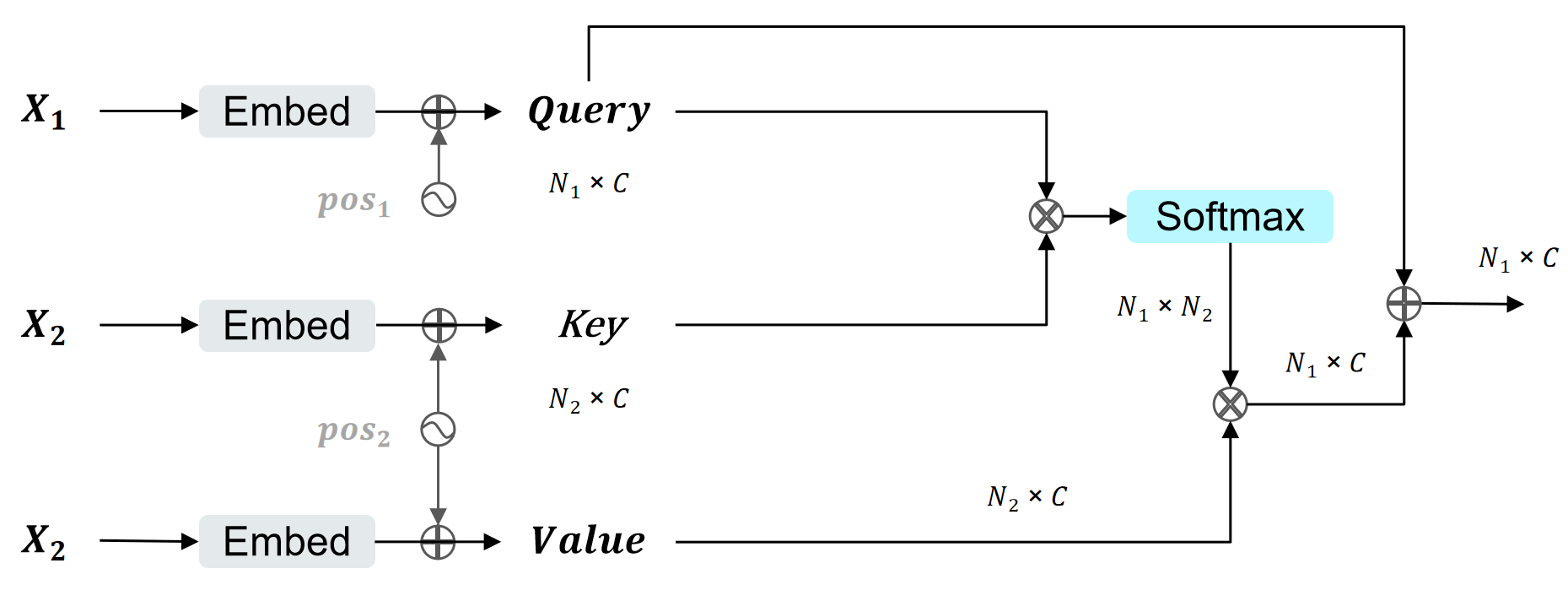}
		\label{fig:4_1}
	}
	\subfigure[Relational cross attention.]{
		\includegraphics[width=0.5\linewidth]{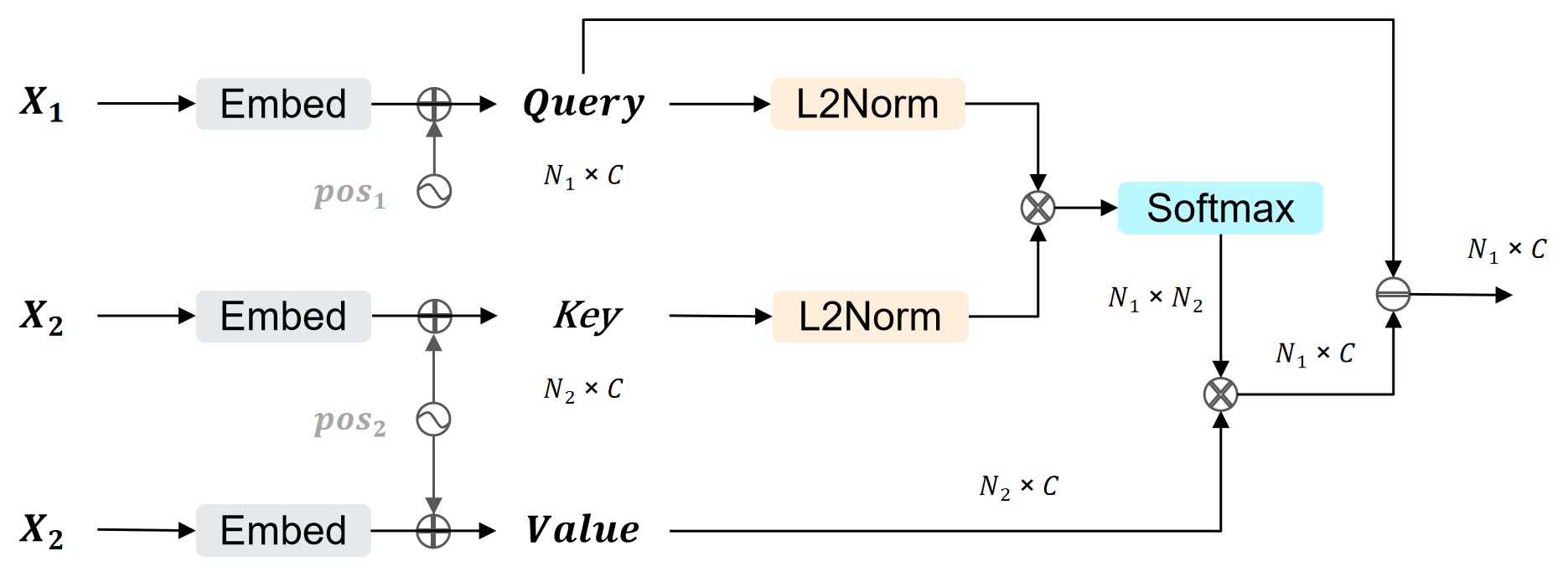}
		\label{fig:4_2}
	}
	\caption{The standard cross attention V.S. the proposed relational cross attention.}
	\setlength{\belowcaptionskip}{-2cm}
	\label{fig:4}
\end{figure}

\begin{equation}
	\label{eq:ca}
	\setlength{\abovedisplayskip}{3pt}
	\setlength{\belowdisplayskip}{3pt}
	Y=Q+softmax(Q\cdot K^{T} )\cdot V
\end{equation}

where $X_{1}$ and $X_{2}$ denote before and after image features, $Q$ denotes $sum(X_{1} ,pos1)$, $K$ denotes $sum(X_{2} ,pos2)$, $V$ denotes $sum(X_{2} ,pos2)$, and $Y$ denotes the output query features. In comparison, our cross attention follows:

\begin{equation}
	\label{eq:oca}
	\setlength{\abovedisplayskip}{3pt}
	\setlength{\belowdisplayskip}{3pt}
	Y=Q-softmax(\frac{Q\cdot K^{T}}{\parallel Q\parallel_{2} \cdot \parallel K\parallel_{2}} )\cdot V
\end{equation}

where $\parallel \cdot \parallel_{2}$ denotes the L2 norm. The modified cross attention is expected to better capture the cross-contextual information from $X_{1}$ and $X_{2}$, leading to improved feature $Y$ ($\mathcal{F}  \in \mathbb{R} ^{\frac{H\cdot W}{S^{2} } \times C}$) with enhanced bi-temporal correspondence.

The second component of RCAM is composed of a cross attention module and an FFN. In this component, query features and query positional encoding become learnable, where the input query features are pixel embeddings $\mathcal{F}  \in \mathbb{R} ^{K \times C}$, with the same dimensionality of query positional embeddings. The Key features and Value features are the Feature $Y$, where the fixed sine Spatial positional encoding serves as the positional encoding. The query features after the cross attention model is passed to a simple three-layer FFN to obtain the final output pixel embedding $\mathcal{F}_{pixel} \in \mathbb{R} ^{K \times C}$.

To leverage multi-scale information, we process with multi-scale transformer decoder operations. Specifically, we obtain the pixel embeddings from the first scale ($\frac{H}{32} \times \frac{W}{32}$), and these embeddings serve as the input query features for the second scale ($\frac{H}{16} \times \frac{W}{16}$). Eventually, we are able to obtain multi-scale pixel embeddings, serving as multi-scale supervision. Considering the computational efficiency, we select the first three scales.

\subsection{Features Constrain Module (FCM)}
\label{S:3.4}
\begin{figure}[H]
	\centering
	\vspace{-0.8cm}
	\includegraphics[width=0.5\linewidth]{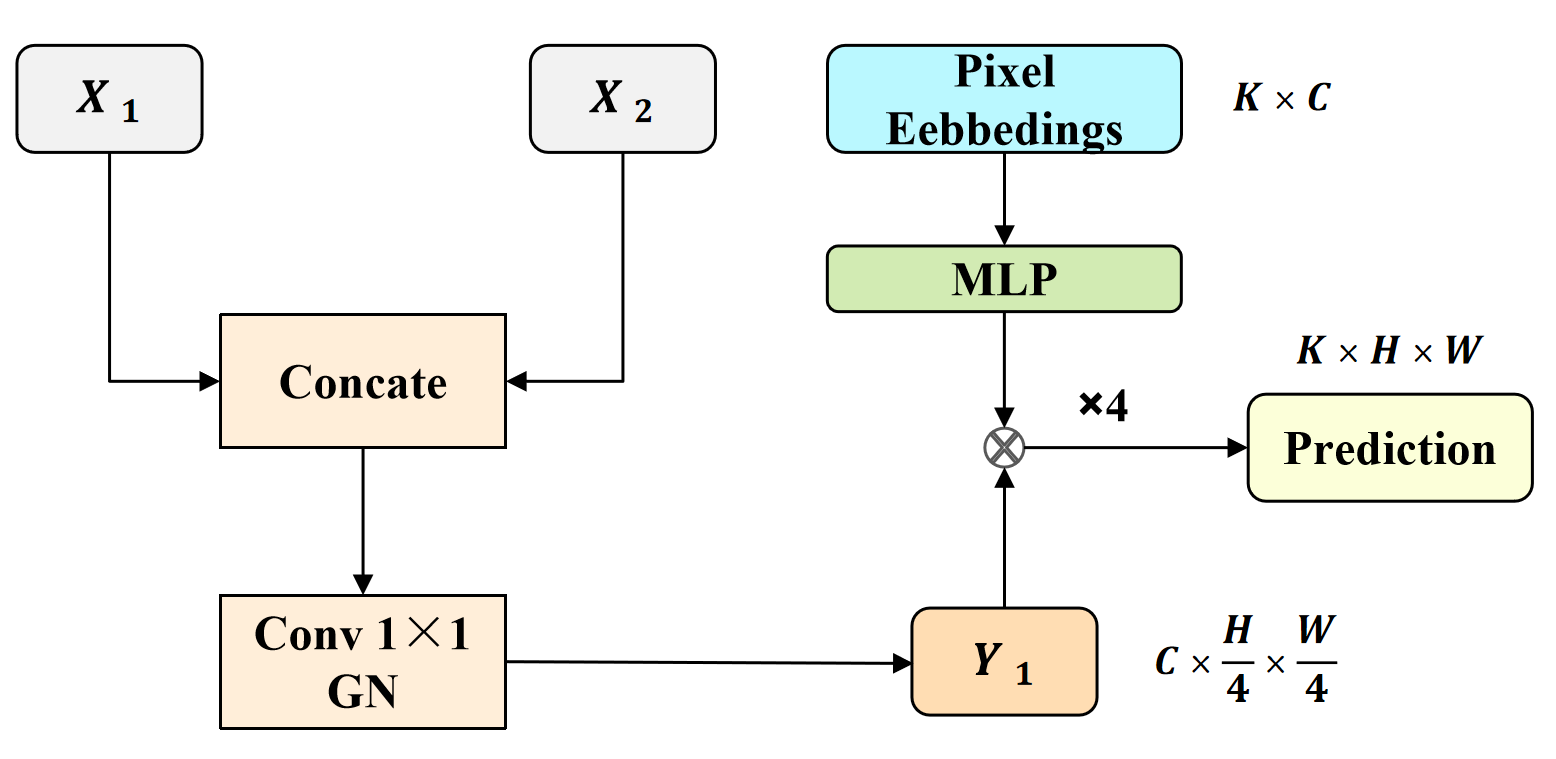}
	\caption{The proposed Features Constrain Module (FCM).}
	\setlength{\belowcaptionskip}{-2cm}
	\label{fig:6}
	\end{figure}

The pixel embeddings from RCAM can be further used to obtain the final change prediction. To refine change detection predictions, We propose a light FCM, as shown in Figure \ref{fig:6}. Specifically, we first utilize a Multi-Layer Perceptron (MLP) with three hidden layers to convert pixel embeddings to segment embeddings $\varepsilon _{segment} \in \mathbb{R}^{K\times C} $. We further concatenate the bi-temporal high-resolution features (1/4) acquired from the Siamese Backbone $\mathcal{F} _{concat} \in \mathbb{R} ^{2C\times \frac{H}{4}\times \frac{W}{4} } $, followed by a $1\times 1$ convolution to achieve projected features $\mathcal{F} _{constrain} \in \mathbb{R} ^{C\times \frac{H}{4}\times \frac{W}{4} } $ ($C=256$), with the same number of channels as the pixel embeddings. The final change map $\mathcal{F}\in \mathbb{R} ^{K\times H\times W} $ is obtained via an upsampling operation ($\times 4$) on the dot product between the segment embeddings $\varepsilon _{segment} $ and the projected features $\mathcal{F}  _{constrain} $ . The proposed FCM is computationally friendly and with improved capability in detecting small objects thanks to the constraints from high-resolution features.

\subsection{Loss Function}
\label{S:3.5}
The proposed RCDT adopts an end-to-end training approach, with each change map prediction supervised by a cross-entropy loss and a dice loss \cite{milletari2016v}. For better supervising multi-scale information and robustness, the designed loss is a summation of multi-scale cross-entropy loss and the dice loss:

\begin{equation}
\label{eq:loss}
\setlength{\abovedisplayskip}{3pt}
\setlength{\belowdisplayskip}{3pt}
\mathcal{L}_{total} =\sum_{l=1}^{N} (\mathcal{L}_{dice}(Pre^{l}, GT^{l})+\alpha \cdot \mathcal{L}_{ce}(Pre^{l}, GT^{l}))
\end{equation}

where $\mathcal{L}_{total}$ denotes the final loss, $N=3$, representing three scales. $Pre^{l}$ and $GT^{l}$ denote the prediction and ground truth at every scale, respectively. $\mathcal{L}_{ce}$ and $\mathcal{L}_{dice}$ denote cross-entropy loss and the dice loss, respectively. $\alpha$ aims to balance the cross-entropy loss and the dice loss, which is empirically set to 0.4.

\section{Experiments}
\label{S:4}
\subsection{Experimental Setup}
\label{S:4.1}

\subsubsection{Datasets}
\label{S:4.1.1}

To explore the capability of our proposed RCDT, we test the performance of RCDT and other competing methods on four popular CD datasets derived from VHR images:
\begin{enumerate}
	\item LEVIR-CD \citep{chen2020spatial} is a widely used building CD dataset that contains 637 very high-resolution (0.5m) Google Earth image pairs with a size of $1024\times 1024$ pixels. We divide LEVIR-CD into training (445), validation (64), and testing (128) sets. To ensure consistency with other methods, we crop image pairs into small patches of size $256\times 256$ without overlap. After cropping, we obtain 7120, 1024, and 2048 pairs of patches for training, validation, and testing sets, respectively.
	\item DSIFN \citep{zhang2020deeply} contains a total of 408 annotated high-resolution (2m) satellite image pairs with a size of $512\times 512$ and with 360 and 48 image pairs for training and testing, respectively. To ensure consistency, we crop image pairs into small patches of size $256\times 256$ without overlap and further separate training and validation sets following a ratio of nine to one. We obtain 5184, 576, and 768 pairs of patches for training, validation, and testing sets, respectively.
	\item CDD \citep{lebedev2018change} contains a total of 16,000 pairs of season-varying remote sensing image patches captured in the same region. Images in CDD are derived from Google Earth with a size of $256\times 256$ and varying spatial resolutions that range from 0.03m to 1m. We divide the CDD dataset into 10,000, 3,000, and 3,000 image pairs for training, validation, and testing, respectively.
	\item SYSU-CD \citep{shi2021deeply} contains remote sensing images with multiple changing types in relatively complex scenarios. It contains a total of 20,000 pairs of aerial image patches with a spatial resolution of 0.5m and a uniform size of $256\times 256$. We divide this dataset into 12,000, 4,000, and 4,000 image pairs for training, validation, and testing.
	\end{enumerate}
	
\subsubsection{Evaluation metrics}
\label{S:4.1.2}
We use $IoU$ (Intersection-over-Union) \citep{everingham2015pascal} and $F1$ score as the main evaluation metrics. These two metrics have been widely used in assessing models’ change detection performances. The $IoU$ and $F1$ scores can be calculated as follows:

\begin{equation}
\label{eq:IoU}
IoU=\frac{TP}{TP+FP+FN}
\end{equation}

\begin{equation}
\label{eq:precision}
precision=\frac{TP}{TP+FP}
\end{equation}

\begin{equation}
\label{eq:recall}
recall=\frac{TP}{TP+FN}
\end{equation}

\begin{equation}
\label{eq:F1}
F1=\frac{2\times precision\times recall}{precision+recall}
\end{equation}

where TP, FP, and FN denote true positive, false positive, and false negative, respectively. Leveraging $precision$ and $recall$, the $F1$ score is an index that measures the overall accuracy of a binary classifier, with 1 being the best value and 0 being the worst.

\subsubsection{Training and testing procedures}
\label{S:4.1.3}

Our method is implemented in Pytorch \citep{paszke2017automatic}. We train RCDT with AdamW \citep{loshchilov2017decoupled} via a poly \citep{chen2017deeplab} learning rate schedule. For all backbones, we set the initial learning rate to 0.0001, weight decay to 0.001, and learning rate multiplier to 0.1. We conduct experiment on three backbones, i.e., R18 \citep{he2016deep}, R50 \citep{he2016deep}, and SwinT \citep{liu2021swin}. Note that all backbones were pre-trained on ImageNet-1K \citep{russakovsky2015imagenet}. We apply a dropout of 0.2 to attention maps and use standard data augmentation techniques for all datasets, including random scale jittering with a factor in [0.5,2.0], random cropping, random horizontal flipping, random color jittering, random noise, and random Gaussian blur. For all datasets, we train all models for 150 epochs, using a crop size of $256\times 256$ and a batch size of 16. We train all models with one RTX TITAN GPU. The testing phase (also conducted on one RTX TITAN GPU with a batch size of one) does not include any data augmentation.

\subsection{Competing state-of-the-art methods}
\label{S:4.2}

To validate the robustness of RCDT, we compare it against several state-of-the-art competing models on the four different datasets mentioned above. We select four convolution-based methods, i.e., IFN \citep{zhang2020deeply}, SNUNET \citep{fang2021snunet}, DSAMNet \citep{shi2021deeply} and FCCDN \citep{chen2022fccdn}, and two transformer-based methods, i.e., BIT \citep{chen2021remote} and ChangeFormer \citep{bandara2022transformer}, as competing models. All models are constructed using their open-sourced codes with the same training strategies as documented in their papers. We give a brief introduction of these competing algorithms below.

\begin{enumerate}
\item IFN \citep{zhang2020deeply} is a deeply supervised image fusion network designed for change detection tasks. IFN extracts bi-temporal features via a Siamese convolutional network and feeds these features to a deeply supervised difference discrimination network(DDN) for change detection.
\item SNUNET \citep{fang2021snunet} is a densely connected Siamese network designed for change detection tasks, which uses the dense skip connection mechanism between the encoder and decoder to capture high-resolution features and fine-grained localization information. In addition, SNUNET involves an Ensemble Channel Attention Module (ECAM) to supervise the learning procedure, aiming to capture the details of representative semantic characteristics.
\item DSAMNet \citep{shi2021deeply} is a deeply supervised (DS) attention metric-based network that employs DS layers to obtain fine-grained bi-temporal features. DSAMNet is featured by its convolutional block attention modules for better feature distinguishment capability.
\item FCCDN \citep{chen2022fccdn} is a feature constraint Siamese network designed for change detection. It first extracts features via a dual encoder-decoder network backbone with a nonlocal feature pyramid structure and fuses bi-temporal features via a connection-based feature fusion module. A self-supervised learning-based strategy is implemented to enhance feature learning capability of the model.
\item BIT \citep{chen2021remote} adopts a Siamese Transformer encoder-decoder structure to enhance bi-temporal feature expression and uses a difference operation to fuse bi-temporal features.
\item ChangeFormer \citep{bandara2022transformer} uses a Siamese Hierarchical Transformer as encoder and an FFN as decoder, where bi-temporal features from each phase of the encoder are fused via the difference operation. The hierarchical fused features are further concatenated and fed to the decoder.
\end{enumerate}

\section{Results}
\label{S:5}

\subsection{Quantitative assessment}
\label{S:5.1}

Table \ref{Table:1} reports the results of the quantitative assessment of the proposed RCDT and other competing models on the testing subsets from four different change detection datasets. To reflect their capabilities, we select $precision$, $recall$, $IoU$, and $F1$ as evaluation metrics. Overall, the proposed RCDT model ranks in the top two in all four datasets, which demonstrates its generalizability and robustness. Specifically, on the LEVIR-CD dataset, relying on the same CNN backbone, i.e., R18 \cite{he2016deep}, the proposed RCDT outperforms ChangeFormer \cite{bandara2022transformer}, transformer-based CD method by 1.71 in $IoU$ and 1.01 in $F1$ score, while having fewer number of parameters. Relying on the vision transformer backbone SwinT \cite{liu2021swin}, the proposed RCDT achieves 85.50 in $IoU$ and 92.18 in $F1$ Score. Note that, however, the proposed RCDT is outperformed by state-of-the-art FCCDN \cite{chen2022fccdn}, with 0.19 lower in $IoU$ and 0.11 lower in $F1$ Score. In addition, from Table \ref{Table:1}, we observe that the proposed RCDT owns relatively high $recall$. For example, the RCDT-R50 achieves the highest $recall$ of 93.55. On the DSIFN dataset, RCDT-SwinT achieves the second-best performance, slightly lower than ChangeFormer \cite{bandara2022transformer}. On the CDD and SYSU-CD datasets, relying on the backbone of SwinT \cite{liu2021swin}, RCDT-SwinT achieves the best performance: $IoU$ of 93.79 and $F1$ Score of 96.80 on the CDD dataset, surpassing the second-best ChangeFormer \cite{bandara2022transformer} by 4.70 in $IoU$ and 2.57 in $F1$ Score; $IoU$ of 67.46 and $F1$ Score of 80.57 on the SYSU-CD dataset, surpassing the second-best BIT \cite{chen2021remote} by 1.15 in $IoU$ and 0.83 in $F1$ Score. In addition, we evaluate the model efficiency from the number of parameters and FLOPs (floating point of operations). From Table \ref{Table:1}, we observe that the proposed RCDT achieves a great trade-off between its performance and model complexity.

From the above quantitative comparison on four different datasets, we notice that our proposed RCDT with SwinT \cite{liu2021swin} backbone (i.e., RCDT-SwinT) achieves satisfactory performance, well-demonstrating its great robustness, efficiency, and generalizability. Some interesting patterns are revealed by investigating RCDT’s performance on these datasets. In LEVIR-CD and DSIFN datasets, where the proposed RCDT achieves the second-best performance, the major change objects are buildings, with limited training samples, i.e., 7,120 in LEVIR-CD and 5,184 in DSIFN. In comparison, changes in CDD and SYSU-CD datasets include a variety of objects, including buildings, roads, forests, grassland, ships, and cars, to list a few. The training sample size for CDD and SYSU-CD reach 10,000 and 13,000. We believe the superior performance of RCDT, specifically on CDD and SYSU-CD, is due to the larger training samples and more diverse change objects, which render a considerably larger vocabulary, thus leading to wider supervision for the proposed Offset Cross Attention Module. In addition, the proposed RCDT model owns high Recall, meaning that it tends to produce very few false negatives (FNs) and is stricter in terms of the changing criteria.
\begin{table}[ht]
	\centering
	\renewcommand\arraystretch{2}
	\caption{Model performance of $Precision$, $Recall$, $IoU$, and $F1$ on datasets of LEVIR-CD, DSIFN, CDD, and SYSU-CD. Note. Pre.: $Precision$; Rec.: $Recall$; FLOPs (floating point of operations) are computed for a given crop size of 256×256. Values in red represent the best performance while values in blue represent the second-best performance. }
	\resizebox{\textwidth}{!}{
		\begin{tabular}{clcccccc}
		\toprule[1.5pt]
		\multicolumn{2}{l}{}             & \textbf{LEVIR-CD}                                       & \textbf{DSIFN}                                          & \textbf{CDD}                                            & \textbf{SYSU-CD}                                        & \textbf{}          & \textbf{}      \\
		\multicolumn{2}{c}{\textbf{}}    & \textbf{Pre. / Rec. / IoU / F1}                         & \textbf{Pre. / Rec. / IoU / F1}                         & \textbf{Pre. / Rec. / IoU / F1}                         & \textbf{Pre. / Rec. / IoU / F1}                         & \textbf{\#params.} & \textbf{FLOPs} \\
		\midrule[0.4pt]
		\multicolumn{2}{c}{IFN}          & \textcolor{blue}{\textbf{92.37}}/88.55/82.51/90.42 & 67.86/53.94/42.96/60.10                                 & 95.35/90.19/86.39/92.70                                 & 76.89/73.11/59.94/74.95                                 & 50.71M             & 41.18G         \\
		\multicolumn{2}{c}{SNUNet}       & 91.62/89.85/83.03/90.73                                 & 60.60/72.89/49.45/66.18                                 & 77.17/63.64/53.55/69.75                                 & 77.75/79.98/65.08/78.85                                 & 12.03M             & 27.44G         \\
		\multicolumn{2}{c}{DSAMNet}      & 92.22/88.71/82.54/90.43                                 & 59.07/69.66/46.98/63.93                                 & 94.54/92.77/88.13/93.69                                 & 74.81/81.86/64.18/78.18                                 & 16.2M              & --             \\
		\multicolumn{2}{c}{FCCDN}        & \textcolor{red}{\textbf{93.07}}/92.52/\textcolor{red}{\textbf{85.69}}/\textcolor{red}{\textbf{92.29}} & 63.62/66.84/48.36/65.19                                 & 94.49/89.63/85.18/92.00                                 & \textcolor{red}{\textbf{83.67}}/75.43/65.76/79.34 & 24.2M              & --             \\
		\multicolumn{2}{c}{BIT}          & 89.24/89.37/80.68/89.31                                 & 68.36/70.18/52.97/69.26                                 & 88.97/82.73/75.03/85.74                                 & 83.03/76.70/66.31/79.74                                 & 3.55M              & 4.35G          \\
		\multicolumn{2}{c}{ChangeFormer} & 92.05/88.80/82.48/90.40                                 & \textcolor{red}{\textbf{88.48}}/\textcolor{blue}{\textbf{84.94}}/\textcolor{red}{\textbf{76.48}}/\textcolor{red}{\textbf{86.67}} & 94.50/93.52/89.09/94.23                                 & 82.73/74.75/64.65/78.53                                 & 117M               & --             \\
		\midrule[0.4pt]
		\multicolumn{2}{c}{RCDT-R18}     & 90.19/92.67/84.19/91.41                                 & 76.11/80.36/64.17/78.17                                 & 93.74/94.12/88.56/93.93                                 & 74.31/80.29/62.84/77.18                                 & 8.47M              & 8.95G          \\
		\multicolumn{2}{c}{RCDT-R50}     & 90.59/\textcolor{red}{\textbf{93.55}}/85.26/92.05       & 76.38/\textcolor{red}{\textbf{88.03}}/69.19/81.79                                 & \textcolor{blue}{\textbf{96.18}}/\textcolor{blue}{\textbf{96.32}}/\textcolor{blue}{\textbf{92.78}}/\textcolor{blue}{\textbf{96.25}} & 75.41/\textcolor{blue}{\textbf{84.80}}/\textcolor{blue}{\textbf{66.43}}/\textcolor{blue}{\textbf{79.83}}                                & 21.41M             & 17.30G         \\
		\multicolumn{2}{c}{RCDT-SwinT}   & 91.12/\textcolor{blue}{\textbf{93.27}}/\textcolor{blue}{\textbf{85.50}}/\textcolor{blue}{\textbf{92.18}}                                 & \textcolor{blue}{\textbf{80.64}}/83.98/\textcolor{blue}{\textbf{69.89}}/\textcolor{blue}{\textbf{82.28}} & \textcolor{red}{\textbf{96.63}}/\textcolor{red}{\textbf{96.97}}/\textcolor{red}{\textbf{93.79}}/\textcolor{red}{\textbf{96.80}} & 75.62/\textcolor{red}{\textbf{86.21}}/\textcolor{red}{\textbf{67.46}}/\textcolor{red}{\textbf{80.57}}                                 & 33.49M             & 23.52G         \\
		\bottomrule[1.5pt]
		\end{tabular}}
		\label{Table:1}    
	\end{table}

\subsection{Qualitative assessment}
\label{S:5.2}

\begin{figure}[]
	\centering
	%\vspace{-0.8cm}
	\includegraphics[width=1\linewidth]{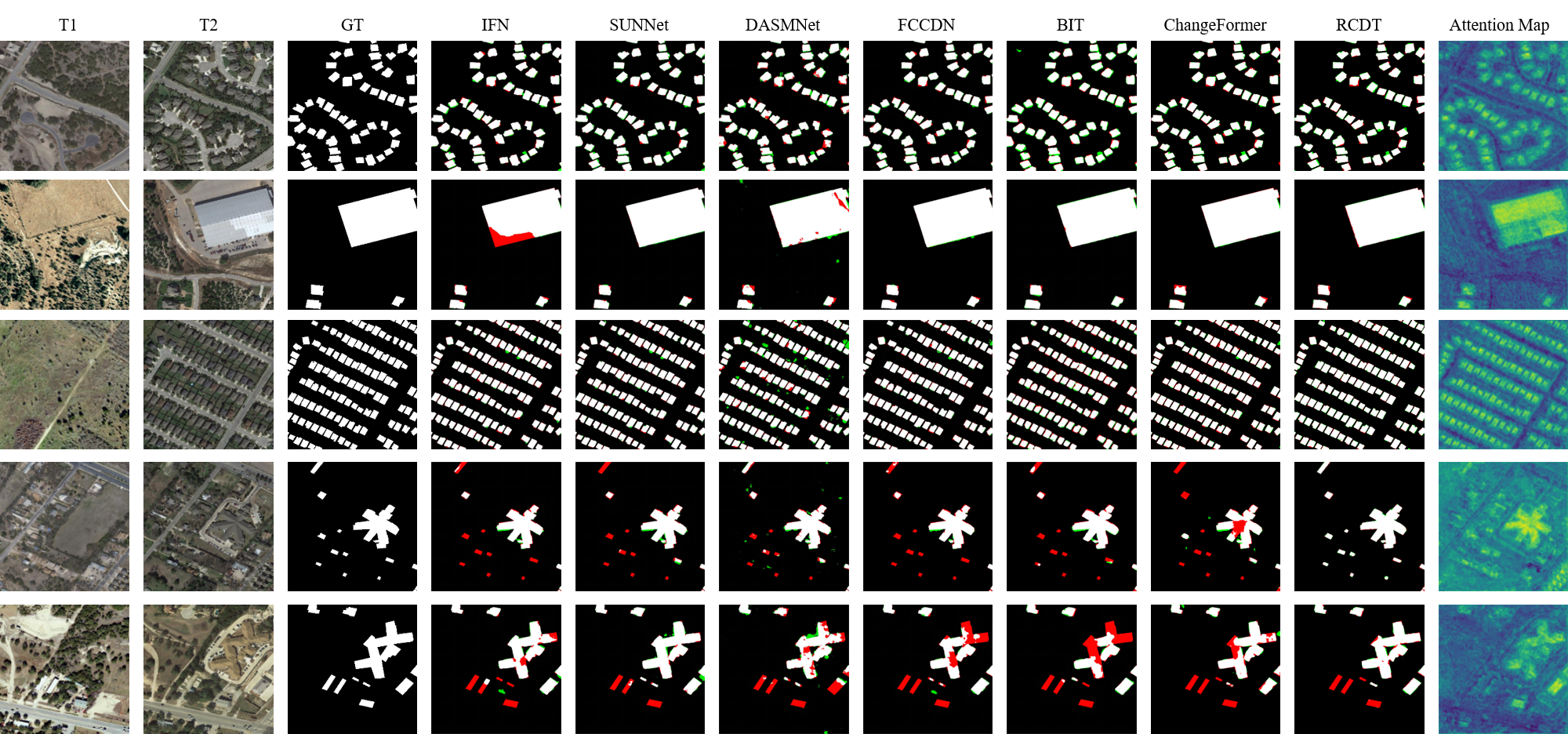}
	\caption{Selected change detection results on the LEVIR-CD dataset. False negative areas are highlighted in red, while false positive areas in green. The last column presents the offset cross attention maps of the proposed RCDT model.}
	\setlength{\belowcaptionskip}{-2cm}
	\label{fig:7}
	\end{figure}

\begin{figure}[]
	\centering
	%\vspace{-0.8cm}
	\includegraphics[width=1\linewidth]{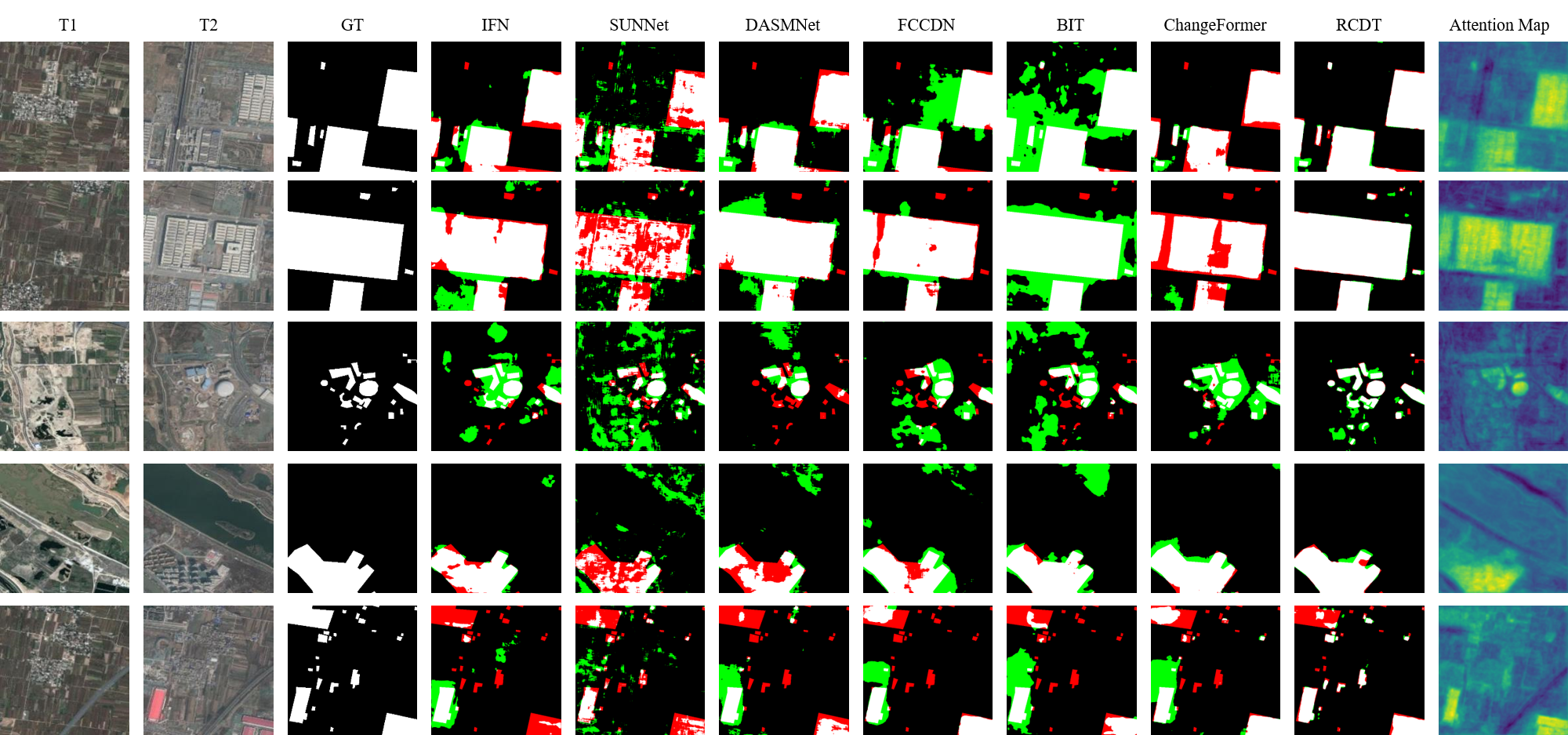}
	\caption{Selected change detection results on the DSIFN dataset. False negative areas are highlighted in red, while false positive areas in green. The last column presents the offset cross attention maps of the proposed RCDT model.}
	\setlength{\belowcaptionskip}{-2cm}
	\label{fig:8}
	\end{figure}

\begin{figure}[]
	\centering
	%\vspace{-0.8cm}
	\includegraphics[width=1\linewidth]{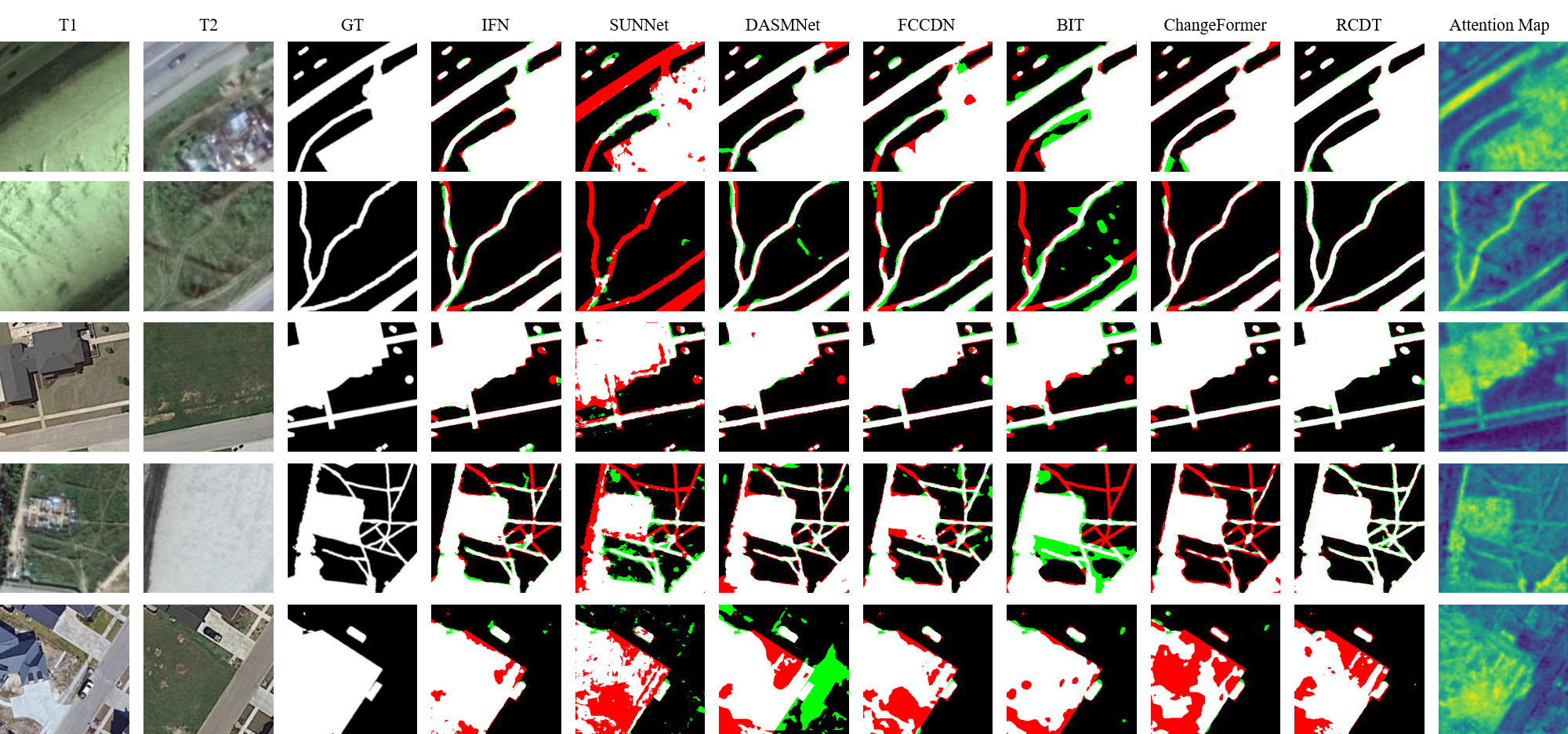}
	\caption{Selected change detection results on the CDD dataset. False negative areas are highlighted in red, while false positive areas in green. The last column presents the offset cross attention maps of the proposed RCDT model.}
	\setlength{\belowcaptionskip}{-2cm}
	\label{fig:9}
	\end{figure}

\begin{figure}[]
	\centering
	%\vspace{-0.8cm}
	\includegraphics[width=1\linewidth]{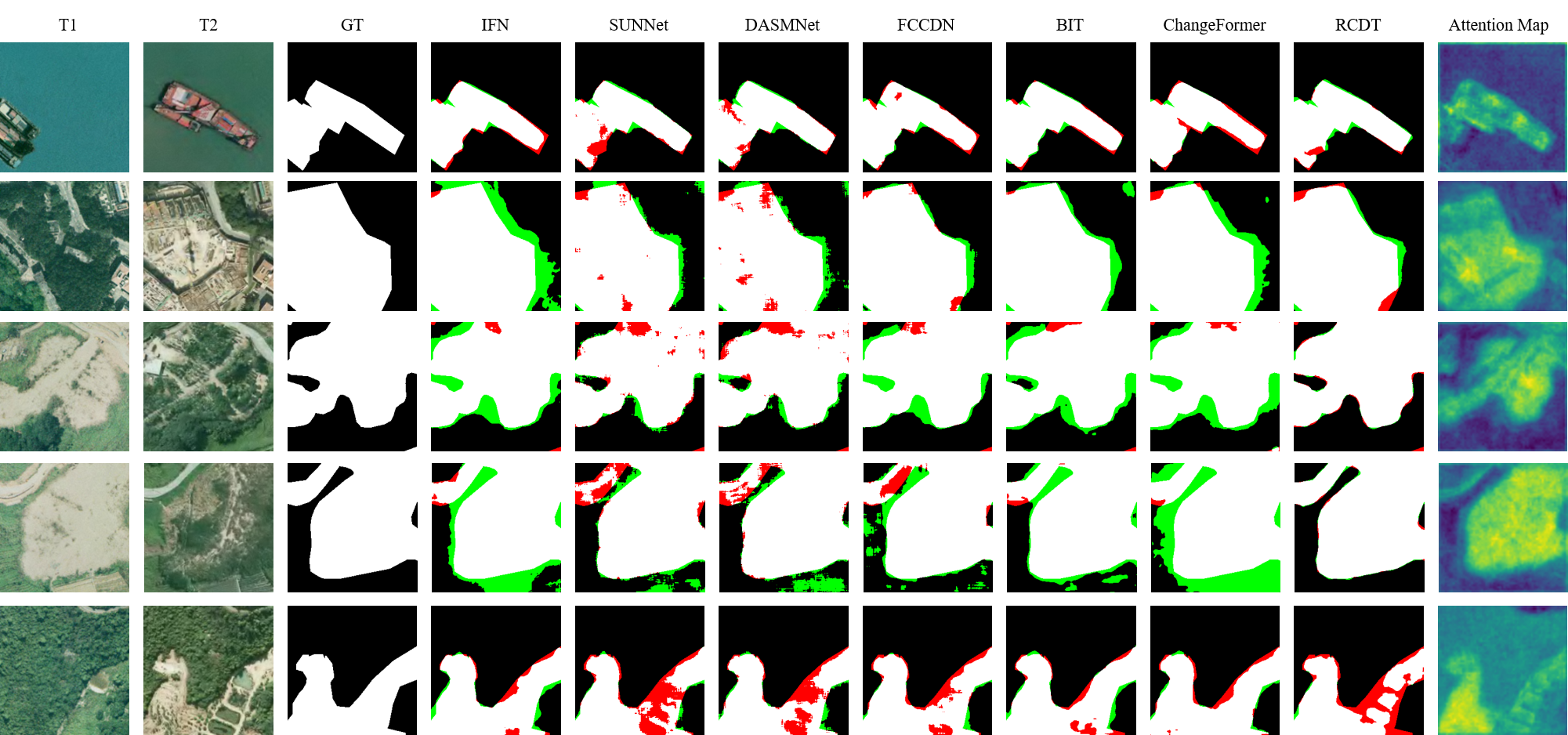}
	\caption{Selected change detection results on the SYSU-CD dataset. False negative areas are highlighted in red, while false positive areas in green. The last column presents the offset cross attention maps of the proposed RCDT model.}
	\setlength{\belowcaptionskip}{-2cm}
	\label{fig:10}
	\end{figure}

We visualize the results from RCDT-SwinT and other six competing models on four CD datasets: LEVIR-CD (Figure \ref{fig:7}), DSIFN (Figure \ref{fig:8}), CDD (Figure \ref{fig:9}), and SYSU-CD (Figure \ref{fig:10}). Among these Figures, the first two columns present the before and after images of the same location, respectively. The third column presents the ground-truthing changes. The fourth to the tenth columns present the detections from IFN \citep{zhang2020deeply}, SNUNET \citep{fang2021snunet}, DSAMNet \citep{shi2021deeply}, FCCDN \citep{chen2022fccdn}, BIT \citep{chen2021remote}, ChangeFormer \citep{bandara2022transformer}, and the proposed RCDT (relying on the SwinT \citep{liu2021swin} backbone). To reveal the capabilities of involved models, we highlight false negative areas in red and false positive areas in green. The eleventh column presents the offset cross attention maps in RCDT. The lighter the color, the higher the attention value. 

For the LEVIR-CD dataset (Figure \ref{fig:7}), all models present satisfactory performance due to the dominance of buildings as the major change object. The proposed RCDT presents superior performance for small buildings, thanks to its many multi-scale strategies that include the FPN \citep{lin2017feature} structure, multi-scale cross attention modules, and multi-scale loss supervision. The DSIFN dataset is also a building-dominant dataset. However, it presents a larger variance in terms of building characteristics compared to the LEVIR-CD dataset, thus leading to the overestimation of changed areas. As shown in Figure \ref{fig:8}, many algorithms tend to produce false positives (in green), while the proposed RCDT resolves this issue thanks to its capability to better harness the bi-temporal features. The CDD dataset includes a variety of changing objects. From Figure \ref{fig:9}, we observe that the competing methods present great performance in detection changes for buildings but weak performance in roads. In general, the detection of the road changes demands models to pay attention to spatially explicit details. However, competing methods do not present such capability due to their insufficiency in harnessing bi-temporal features. For convolution-based methods, given their limited field of view, they fail to obtain global, long-range semantic information, thus leading to incontinence of road detections. For the competing transformer-based methods, despite that they address the short-range issue in bi-temporal features, the inner relations of bi-temporal features are not explicitly explored, presumably leading to the failure of detecting subtle road changes. Relying on the proposed RCAM, our RCDT is capable of detecting obvious building changes as well as subtle road changes. As shown in Figure \ref{fig:10}, for the SYSU-CD dataset with various changing categories (e.g., buildings, roads, vegetation, ships, etc.), RCDT is sensitive to a variety of changing objects, evidenced by its produced finer and more accurate boundaries. These results prove that RCDT achieves high-resolution expression via FCM and is able to produce spatially explicit change detection results.
	
\subsection{Ablation studies}
\label{S:5.3}
To explore the impact of model components, we conduct a series of ablation experiments with the proposed RCDT model relying on SwinT \citep{liu2021swin} as the backbone in the LEVIR-CD dataset.

\textbf{Deep supervise.} We first investigate the multi-scale loss. As shown in Table \ref{Table:2}, the model without the cross-entropy loss and without dice loss of the last scale achieves 85.39 $IoU$. The addition of multi-scale cross-entropy loss or dice loss further improves model performance, suggesting that the multi-scale loss is able to supervise multi-scale change features. The involvement of both cross-entropy loss and dice loss achieves the best performance with 85.50 in $IoU$.

\begin{table}[H]
\centering
\caption{The impact of multi-scale cross-entropy loss ($\mathcal{L} _{ce}$) and dice loss ($\mathcal{L} _{dice}$) on model performance. }
    \begin{tabular}{c|c|c}
    $\mathcal{L} _{ce}$ deep supervise        & $\mathcal{L} _{dice}$ deep supervise        & $IoU$           \\ \midrule[1.5pt]
    \multicolumn{1}{l|}{} & \multicolumn{1}{l|}{} & 85.39         \\
    \checkmark            &                       & 85.43         \\
                          & \checkmark            & 85.45         \\
	\checkmark            & \checkmark            & \textbf{85.50}
    \end{tabular}
    \label{Table:2}   
\end{table}

\begin{figure}[H]
    \centering
    \includegraphics[width=0.5\linewidth]{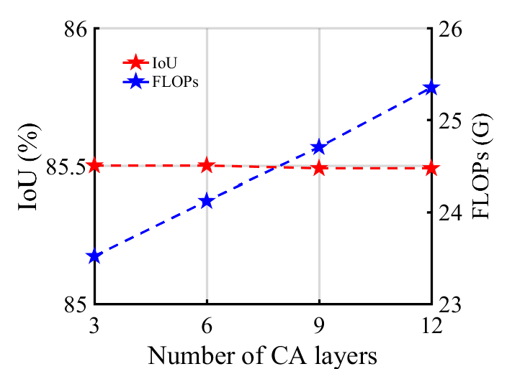}
    \caption{The impact of the number of cross attention layers on RCDT’s performance of efficiency.}
    \setlength{\belowcaptionskip}{-2cm}
	\label{fig:11}
    \end{figure}

\textbf{The number of cross attention layers.} Figure \ref{fig:11} reports the results of RCDT trained with a different number of cross attention layers. Considering the multi-scale design, the number of cross attention layers is often set as the multiples of three. In this experiment, we select 3, 6, 9, and 12. The results suggest that the proposed RCDT with three cross attention layers achieves the best performance while saving additional computation costs introduced by the excessive cross attention layers.

\begin{figure}[H]
    \centering
    \includegraphics[width=0.5\linewidth]{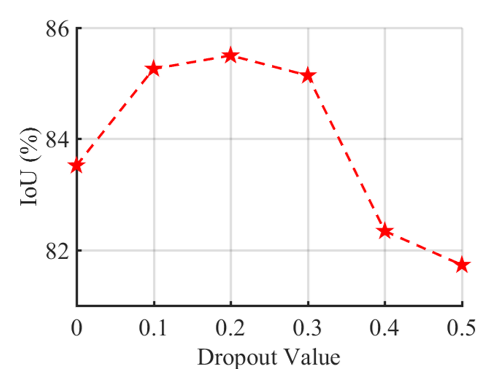}
    \caption{The impact of different dropout values for attention maps on model performance.}
    \setlength{\belowcaptionskip}{-2cm}
	\label{fig:12}
    \end{figure}

\textbf{Dropout value for attention maps.} Figure \ref{fig:12} shows RCDT’s performance along with different dropout values of attention maps. Given the limited training sample size, our RCDT avoids overfitting by implementing the dropout mechanism. The results suggest that our RCDT reaches the best performance when the dropout is set to 0.2, outperforming the default 0.1 by 0.24 $IoU$. When the dropout continues to increase above 0.2, the model performance reduces.

\textbf{Offset CA or Standard CA.} Compared with standard cross attention, offset cross attention is different from two perspectives, i.e., the utilization of cosine similarity to replace dot product when dealing with query and key features and the utilization of subtraction instead of addition to obtain the final attention map. Table \ref{Table:3} shows that both cosine similarity and subtraction are able to improve the model’s performance, especially the cosine similarity. The above results suggest that offset cross attention benefits the learning of bi-temporal correspondence, thanks to its unique cosine similarity to process query and key features and the subtraction operation to obtain the attention map.
\begin{table}[H]
    \centering
    \caption{The impact of cosine similarity and subtraction in the offset cross attention on model’s performance.} 
    \begin{tabular}{c|c|c}
    cosine similarity & subtraction & $IoU$            \\ \midrule[1.5pt]
                      &             & 84.63          \\
	\checkmark                 &             & 85.21          \\
                      & \checkmark           & 84.65          \\
	\checkmark        & \checkmark           & \textbf{85.50}
    \end{tabular}
    \label{Table:3}
    \end{table}

\textbf{The impact of FFN and self-attention mechanism.} FFN is a multi-layer perceptron in nature, with similar functionality as a $1\times 1$ convolution. The self-attention mechanism aims to enhance the information expression in single-phase images. Here, we validate the importance of these two components. As shown in Table \ref{Table:4}, when the Relational Cross Attention Module (RCAM) removes FFN, the $IoU$ drops 2.36, highlighting the importance of FFN in model performance. However, when the proposed RCAM involves the self-attention mechanism, the performance of the model does not improve but leads to additional computational burdens. It reveals that the self-attention mechanism is not necessary for the proposed RCDT model. 
\begin{table}[H]
    \centering
    \caption{The impact of FFN and self-attention mechanism on model performance. Note. OCA: offset cross attention; CA: standard cross attention; SA: self attention; FFN: forward feed network} 
    \begin{tabular}{lc|l|c}
                                          &               & \multicolumn{1}{c|}{IoU}            & FLOPs                \\ \midrule[1.5pt]
    \multicolumn{1}{l|}{RCDT(ours)}       & OCA-CA-FFN    & 85.50                                & 23.52G               \\
    \multicolumn{1}{l|}{- FFN}            & OCA-CA        & $83.14_{(-2.36)}$                        & 23.29G               \\
    \multicolumn{1}{l|}{+ Self Attention} & OCA-CA-SA-FFN & $85.50_{(+0.00)}$                        & 24.18G               
    \end{tabular}
    \label{Table:4}    
\end{table}

\textbf{The impact of Features Constrain Module (FCM).} The FCM aims to provide high-resolution constraints. Its lightweight design does not introduce much computational burden. In this experiment, we remove FCM, i.e., directly feeding the features from the offset cross attention to a prediction head that consists of $1\times 1$ convolution. As shown in Table \ref{Table:5}, we notice that the removal of FCM leads to reduced $IoU$ by 1.41 and FLOPs by 1.35. 
\begin{table}[H]
    \centering
    \caption{The impact of Features Constrain Module (FCM) on model performance.} 
    \begin{tabular}{l|l|c|}
               & \multicolumn{1}{c|}{IoU} & FLOPs  \\ \midrule[1.5pt]
    RCDT(ours) & 85.50                     & 23.52G \\
    - FCM      & $84.09_{(-1.41)}$            & 22.17G
    \end{tabular}
    \label{Table:5}    
\end{table}

\section{Discussion}
\label{S:6}

\subsection{The cross attention mechanism benefits remote sensing change detection tasks}
\label{S:6.1}

Existing AI-based CD methods largely rely on three functionality-enhancing modules, i.e., semantic enhancement, attention mechanisms, and correspondence enhancement. The stacking of these modules leads to great model complexity. To unify these three modules into a simple pipeline, we introduce the RCDT module that involves the cross attention mechanism. For CD tasks on VHR images, we argue that more attention should be paid to the correspondence of bi-temporal features, besides the semantic segmentation of bi-temporal features. To validate the validity of the cross attention mechanism on VHR image change detection tasks, we conduct experiments on four publically available datasets. The results reveal the superiority of our proposed RCDT compared to other competing models in terms of performance, efficiency, and robustness.

\begin{figure}[h]
    \centering
    \vspace{-0.15in}
    \begin{minipage}{1\linewidth}
        \subfigure[LEVIR-CD]{
            \includegraphics[width=0.48\linewidth]{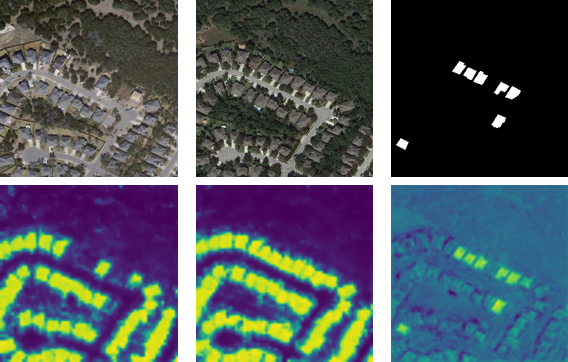}
            \label{fig:13_1}
            }
        \subfigure[DSIFN]{
            \includegraphics[width=0.48\linewidth]{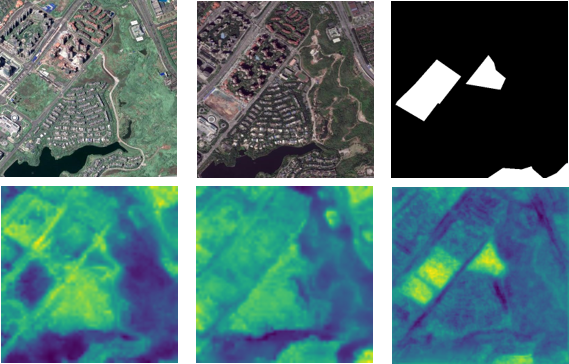}
            \label{fig:13_2}
            }     
    \end{minipage}
    \vskip -0.3cm
    \begin{minipage}{1\linewidth}
        \subfigure[CDD]{
            \includegraphics[width=0.48\linewidth]{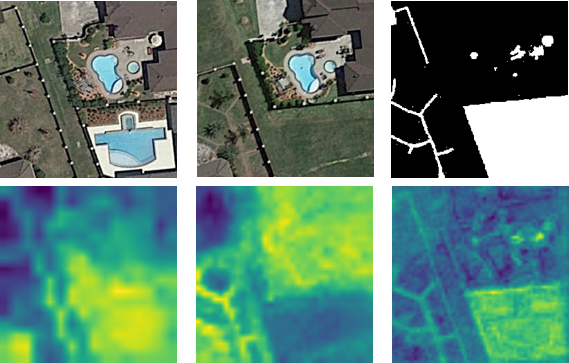}
            \label{fig:13_3}
            }
        \subfigure[SYSU-CD]{
            \includegraphics[width=0.48\linewidth]{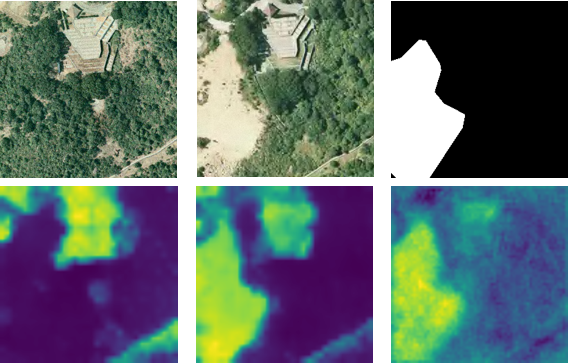}
            \label{fig:13_4}
            }     
    \end{minipage}
    \vspace{-0.18in}
    \caption{The visualization of selected bi-temporal feature maps and the offset cross attention maps. The first row presents the before image, after image, and the ground-truthing changes, respectively. The second row presents the bi-temporal feature maps of before and after images and the offset cross attention map, respectively. The lighter the color, the higher the attention value.}
    \vspace{-0.2in}
    \label{fig:13}
\end{figure}

We visualize selected bi-temporal feature maps and the offset cross attention maps in RCDT to explain how RCDT obtains feature correspondence in Figure \ref{fig:13}, where the first row presents the before image, after image, and the ground-truthing changes, respectively. The second row presents the bi-temporal feature maps of before and after images and the offset cross attention map, respectively. The lighter the color, the higher the attention value.

From the bi-temporal features maps in Figure \ref{fig:13_1}, we observe that the Siamese network successfully extracts buildings’ semantic features from before and after images, evidenced by their clear boundaries. For CD tasks, however, the changes between these building semantic features are all that matter. The offset cross attention map after the cross attention mechanism highlights the potentially changed regions with sharp details of building boundaries, which demonstrates the effectiveness of RCDT in learning the correspondence of bi-temporal features. Besides its capability to capture detailed boundaries for areas with homogenous buildings, the proposed RCDT is able to capture the bi-temporal correspondence for areas with heterogenous buildings with great variance in size and shape (Figure \ref{fig:13_2}). In addition, our approach can effectively extract changes in roads. In Figure \ref{fig:13_3}, the offset cross attention map is able to reflect road changes, although the attention values of roads are not as strong as the ones of buildings at the bottom-right corner. Besides manmade objects, such as buildings and roads, the proposed RCDT is sensitive to natural vegetation. From Figure \ref{fig:13_4}, we notice that the feature map for the before image focuses on the expression of buildings, while the feature map for the after image focuses on the expression of bare lands. After the cross attention mechanism, focuses are directed to the expression of vegetation-bare land changes. 

From the above observations, we conclude that the cross attention mechanism is characterized by its simplicity and effectiveness, as it can capture long-range semantic context, attention to changed areas, and accurate correspondence of bi-temporal features via s simple cross attention structure. The proposed RCDT with the cross attention mechanism presents great generalizability and robustness, evidenced by its superior performance in capturing changes in complex scenes with a variety of objects (e.g., buildings, roads, vegetation, bare land, etc.). These pieces of evidence suggest that the cross attention mechanism can greatly benefit remote sensing CD tasks.

\subsection{Limitations and future pathways of remote sensing change detection}
\label{S:6.2}

The involvement of Transformer structures in obtaining global semantic information is not new, as numerous efforts, e.g., \citep{chen2021remote,bandara2022transformer,liu2022cnn,shi2022divided}, have been made to incorporate Transformer structures and achieved great results. Our study is built upon these efforts by explicitly introducing the cross attention mechanism in our proposed RCDT. Several limitations of this study need to be acknowledged. Relying on the designed multi-scale training strategy, great performance improvement has been noted for our RCDT. However, we also observe that RCDT fails to, in certain cases, fully harness the multi-scale information, especially for tiny objects. A potential resolution for this issue could be the development of a novel pyramid structure or a loss function specifically for tiny objects. Second, the four publically available datasets we used in our study contain images with a spatial resolution ranging from 0.03 m to 2 m. More experiments are still needed to evaluate the performance of our model for image resolution beyond this range (both finer and coarser). Finally, when CD datasets contain insufficient or homogenous samples, the proposed model may fail to capture robust bi-temporal feature corresponds, thus leading to overfitting. We believe designing novel CD data augmentation approaches can mitigate this issue.

Our ultimate goal is to design a “Cross Attention is All You Need” simple structure for all remote sensing detection tasks. In real cases, however, the computational demand is huge for the attention calculation of high-resolution features. Thus, we implement FPN and FCM modules for model simplification without notably sacrificing accuracy. We encourage more efforts to be made to develop models that rely completely on cross attention structures for remote sensing CD tasks.

\section{Conclusion}
\label{S:7}

Remote sensing change detection has always been challenging task. Existing AI-based change detection (CD) methods tend to rely on emantic enhancement, correspondence enhancement, and spatial or channel attention mechanism, whic makes these models complicated, computationally demanding, and difficult to converge. We believe it is necessary to simplify existing remote sensing CD pipelines without sacrificing the change detection accuracy. 

In this work, we propose Relational Change Detection Transformer (RCDT), a novel change detection paradigm that achieves efficient and highly accurate CD for very high resolution (VHR) images. The proposed RCDT consists of three major components, a weight-sharing Siamese Backbone to obtain bi-temporal features, a Relational Cross Attention Module (RCAM) that implements offset cross attention to obtain bi-temporal relation-aware features, and a Features Constrain Module (FCM) to achieve the final refined predictions with high-resolution constraints.

Through extensive experiments on four different publically available datasets, i.e., LEVIR-CD, DSIFN, CDD, and SYSU-CD, our proposed RCDT exhibits superior CD performance over other competing methods. The implemented cross attention mechanism has been proved to facilitate the acquisition of bi-temporal feature correspondence, which is responsible for its great generalizability and efficiency. The proposed RCDT, with satisfactory tradeoff between its performance and complexity, achieves the second place on LEVIR-CD tests (85.50 $IoU$) and DSIFN tests (69.89 $IoU$), and the first place on CDD tests (93.79 $IoU$) and SYSU-CD tests (67.46 $IoU$). Our study indicates the great potential of cross attention mechanism in capturing correspdence between bi-temporal images. We encourage future remote sensing CD efforts to be made towards this direction.

\bibliographystyle{unsrtnat}
\bibliography{RCDT}  %%% Uncomment this line and comment out the ``thebibliography'' section below to use the external .bib file (using bibtex) .





% \end{thebibliography}

\end{document}